\documentclass[sigconf]{acmart}

\usepackage[utf8]{inputenc} 
\usepackage[T1]{fontenc}    
\usepackage{hyperref}       
\usepackage{url}            
\usepackage{booktabs}       
\usepackage{amsfonts}       
\usepackage{nicefrac}       
\usepackage{microtype}      
\usepackage{xcolor}         
\usepackage{color}
\usepackage{cases}

\usepackage{makecell}
\usepackage{multirow}
\usepackage{microtype}
\usepackage{booktabs}
\usepackage[utf8]{inputenc}
\usepackage{listings}
\usepackage{caption}
\usepackage{dsfont}
\usepackage{graphicx}
\usepackage[linesnumbered,ruled,lined]{algorithm2e}
\usepackage{subfig}
\usepackage{ulem}
\usepackage{afterpage}
\usepackage{xspace}
\usepackage{adjustbox}
\usepackage{paralist}
\usepackage{fancybox}

\usepackage[textsize=tiny]{todonotes}

\newcommand{\method}{{GraphStorm}\xspace}

\newtheorem{thm:eg}{Example}

\def \P {\mathbf{P}}

\def \S {\mathcal{S}}

\newcommand{\hide}[1]{}
\newcommand{\easy}{Easy to use\xspace}
\newcommand{\scalable}{Scalable\xspace}
\newcommand{\expert}{Expert-friendly\xspace}

\newcommand{\battleTested}{In production\xspace}
\newcommand{\myEmphasis}[1]{{\bf{ #1}}}

\AtBeginDocument{%
  \providecommand\BibTeX{{%
    \normalfont B\kern-0.5em{\scshape i\kern-0.25em b}\kern-0.8em\TeX}}}

\copyrightyear{2024}
\acmYear{2024}
\setcopyright{acmlicensed}\acmConference[KDD '24]{Proceedings of the 30th ACM SIGKDD Conference on Knowledge Discovery and Data Mining}{August 25--29, 2024}{Barcelona, Spain}
\acmBooktitle{Proceedings of the 30th ACM SIGKDD Conference on Knowledge Discovery and Data Mining (KDD '24), August 25--29, 2024, Barcelona, Spain}
\acmDOI{10.1145/3637528.3671603}
\acmISBN{979-8-4007-0490-1/24/08}

\begin{document}
%

\title{GraphStorm: all-in-one graph machine learning framework for industry applications}
%

%


\author{Da Zheng, Xiang Song, Qi Zhu, Jian Zhang, Theodore Vasiloudis, Runjie Ma, Houyu Zhang, Zichen Wang, Soji Adeshina, Israt Nisa, Alejandro Mottini, Qingjun Cui, Huzefa Rangwala, Belinda Zeng, Christos Faloutsos, George Karypis}
\affiliation{%
  \institution{Amazon}
  \country{USA}\\
  {\{dzzhen, xiangsx, qzhuamzn, jamezhan, thvasilo, runjie, zhanhouy, zichewan,\\
  adesojia, nisisrat, amottini, qingjunc, rhuzefa, zengb, faloutso, gkarypis\}}@amazon.com
}

\renewcommand{\shortauthors}{Da Zheng et al.}

\date{30 July 1999}

\begin{abstract}

Graph machine learning (GML) is effective in many business applications.
However, making GML easy to use and applicable to industry applications
with massive datasets remain challenging.
We developed \method, which provides an end-to-end solution
for scalable graph construction, graph model training and inference. 
\method has the following desirable properties:
(a) \myEmphasis{\easy}: it can perform graph construction and model training
and inference with just a single command;
(b) \myEmphasis{\expert}: \method contains many advanced GML modeling
techniques to handle complex graph data and improve model performance;
(c) \myEmphasis{\scalable}: every component in GraphStorm can operate on
graphs with billions of nodes and can scale model training and
inference to different hardware without changing any code.
GraphStorm has been used and deployed for
over a  {\em dozen} {\em billion-scale} industry applications after
its release in May 2023. It is open-sourced in
Github: \url{https://github.com/awslabs/graphstorm}.

\end{abstract}

\maketitle

\section{Introduction}
Recent research has demonstrated the value of GML across a range of
applications and domains, such as social networks and e-commerce.
However, deploying such GML solutions to solve real business problems remains
challenging for three reasons. First,
industry graphs are massive, usually in the order of many millions or even billions
of nodes and edges. Second, industry graphs are complex. They are usually
heterogeneous with multiple node types and edge types. Some nodes and edges are associated
with diverse features, such as numerical, categorical and text/image features,
while some other nodes or edges have no features. Third, many applications
do not store data in a graph format. To apply GML to these data,
we need to first construct a graph. Defining a graph schema is part of graph modeling and
often requires multiple rounds of trials and errors.

\begin{figure}
    \centering
    \includegraphics[width=\linewidth]{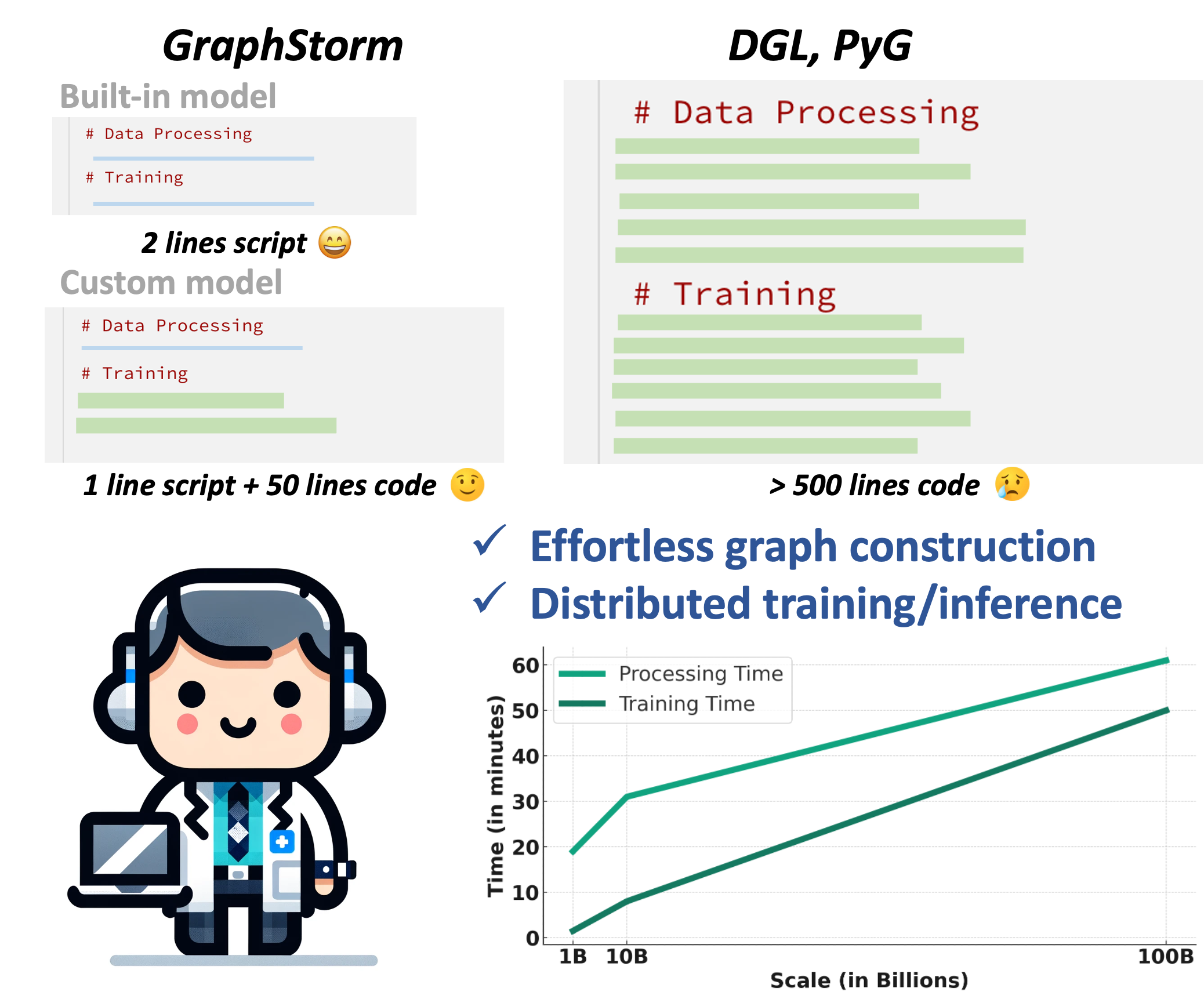}
    \caption{Easy and Scalable GML with GraphStorm.}
    \label{fig:graphstorm_teaser}
    \vspace{-5mm}
\end{figure}

Today users need to manually develop complex code with libraries such as
DGL \cite{dgl} or Pytorch-Geometric \cite{pyg}.
This usually requires users to have significant modeling expertise in GML.
Furthermore, even if a user has the required expertise, it requires
significant efforts to develop techniques to
achieve good performance and scale model training and inference to large graphs
on a cluster of CPU/GPU machines. These challenges prevent GML from becoming
a commonly used technique in industry.

To address these challenges, we developed GraphStorm, which is a no-code/low-code
framework for scientists in the industry to develop, train, and deploy GML models
for solving real business problems. It provides end-to-end pipelines for
graph construction, model training and deployment to simplify every step of
developing GML models.
It balances the usability, flexibility, modeling capability, 
training efficiency and scalability so that users who are not familiar
with GML can easily apply it to their applications. As illustrated in
Figure \ref{fig:graphstorm_teaser}, GraphStorm allows users to develop GML models
and scale them to industry graphs with billions of edges by using
a few lines of code.
To scale to graphs with hundreds of billions of edges, GraphStorm is built on
top of the distributed GNN system DistDGL \cite{distdgl}, adopts
scalable algorithms and provides efficient implementations for these algorithms.
To help users quickly prototype a GML model, GraphStorm
provides a command line interface that allows users to run model training
and inference with a single command.
Once a GML model prototype is ready, GraphStorm provides two options to help
users further improve model performance. It provides many built-in modeling
techniques to improve performance; it supports custom model API to allow users
to develop their own model implementations. GraphStorm allows model training/inference
to run on a single machine and scale it to
a cluster of machines without changing any line of code.

We evaluate the scalability and modeling capability of GraphStorm with two public
graph data with hundreds of millions of nodes and synthetic graph data ranging
from 1 billion edges to 100 billion edges.
GraphStorm is scalable and efficient in processing and modeling graphs with complex features.
It finishes every step needed for training a GML model on billion-scale graphs,
such as MAG and Amazon Review within hours. Even on a synthetic graph with
100 billion edges, we can process the data and train a GNN model with a few
hours.
We further show that GraphStorm provides many practical and effective techniques
to help users improve graph model performance in the benchmark datasets.
For example, GraphStorm allows users to freely try out different
graph schemas and a proper graph schema can improve model performance by up to 15\% on
our benchmark datasets without changing the model architecture and model training
procedure; BERT+GNN fine-tuning can improve model performance on text-rich
graphs by up to 17.6\% in our benchmarks; GNN distillation can improve
the performance of a BERT model by 8\%.
The scalability and practical modeling capablities are essential for industry production.
We have successfully developed multiple GML models with GraphStorm
that outperform production models and are deployed in production.

%

Our contributions are summarized as below:
\begin{itemize}
    \item {\bf \easy}: GraphStorm provides users
    an easy solution from graph construction to GML model training and inference
    with a single command line for real-world applications.
    \item  {\bf \expert}: GraphStorm provides many built-in modeling
    techniques to help scientists 
    improve model performance in complex industry graphs without writing
    any code. It also provides a custom model API for users to develop
    and train their own models.
    \item {\bf \scalable}: GraphStorm ensures that every step for GML model
    development and deployment can scale to graphs with {\em billions} of nodes
    and {\em hundreds of billions} of edges and can scale model training and
    inference in different hardware (e.g., from single GPU to
    multi-machine multi-GPU) without any code modification.
    \item {\bf \battleTested}: GraphStorm has been deployed in production on billion-scale graphs,
    for over 6 months.
\end{itemize}

\section{Related works}

There are many existing GML frameworks. DGL \cite{dgl} and
Pytorch-Geometric \cite{pyg} are two popular frameworks that provide low-level
API for writing graph neural network models (GNNs). Users need to write
complex code to address many practical problems in graph
applications. On top of these low-level GML frameworks, people developed higher-level
frameworks specialized for one type of GML problems. For example,
DGL-KE \cite{dglke} and Pytorch-BigGraph \cite{pbg} contain
a set of knowledge graph embedding models and provide scalable training solutions.
TGL \cite{tgl} provides a set of models for continuous-time
temporal graphs and Pytorch-Geometric Temporal \cite{pygt} contains a set of models for
spatiotemporal models.
These high-level frameworks provide command-line interfaces to train
models without writing code, which simplifies model development.

AliGraph \cite{aligraph}, Euler \cite{euler}, AGL \cite{agl}, PGL \cite{pgl} and
TensorFlow-GNN \cite{tfgnn} are industry GNN frameworks. They are
built for distributed training and inference on large heterogeneous graphs.
AliGraph, Euler and TensorFlow-GNN adds GNN-related functionalities to TensorFlow,
such as graph data structure, message passing operations, mini-batch
sampling on graph data. AGL is built on top of MapReduce for distributed training
and inference of GNN. PGL is a GNN framework built on top of Paddle-Paddle \cite{pgl}.
One of the key design choices is how to perform mini-batch sampling for GNN training.
TensorFlow-GNN and AGL choose the option of preprocessing the input graph to generate
mini-batch graphs and save them to disks, while AliGraph, Euler and PGL choose
on-the-fly sampling. The main benefit of on-the-fly sampling is that
it allows users to easily change some hyperparameters of GNN, such as the number of GNN
layers and fanout, to tune model performance.
GraphStorm is an industry GML framework built on top of DGL and provides
high-level GML functionalities, such as end-to-end training/inference pipelines
and advanced GML techniques. To support fast prototyping, GraphStorm adopts
on-the-fly sampling.

CogDL \cite{cogdl} is a high-level GML library that provides many built-in
GML methods and benchmark datasets. CogDL is built for researchers to
reproduce model performance and set up standard benchmarks for comparing
the performance of different models. GraphStorm is built for scientists to
apply GML to industry applications. Therefore, GraphStorm does not contain built-in
benchmark datasets and instead, provides users an easy way to load tabular data into GraphStorm
for model training and inference. GraphStorm is much more scalable than CogDL because
it intends to address the scalability challenges in the industry.

\section{Design}

\begin{figure*}
\centering
\includegraphics[width=0.9\textwidth]{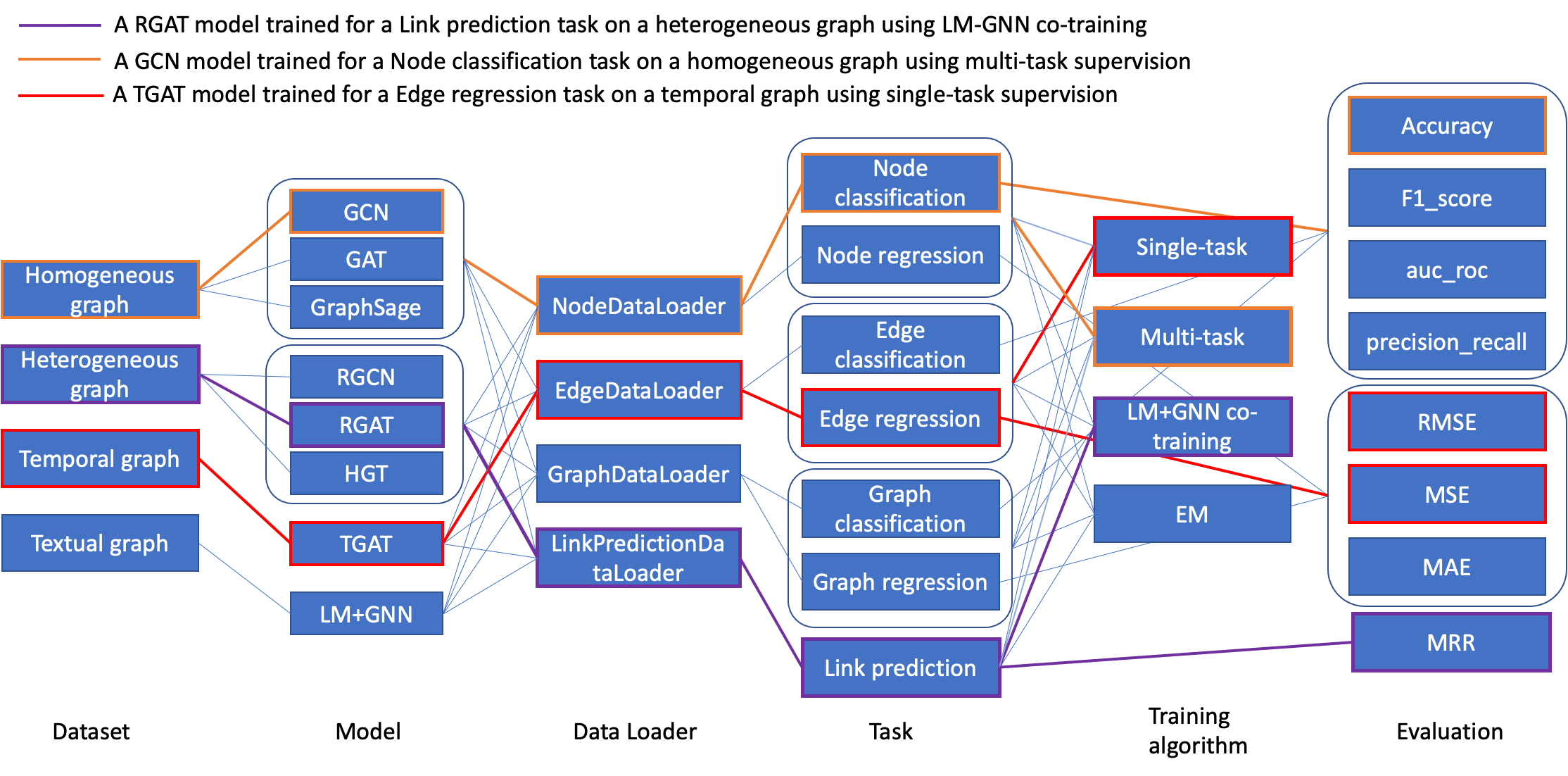}\\
\vspace{-3mm}
\caption{The functionalities of GraphStorm. The colored lines show examples of
constructing a complete model solution in GraphStorm.}
\label{fig:funcs}
\end{figure*}

GraphStorm is designed as a general framework to accelerate GML adoption in
the industry. It provides functionalities to help every step in the journey
of developing and deploying GML models in the industry,
including graph construction, model prototyping, model tuning and model deployment.
Typically, graph modeling starts by defining the right graph schema.
GraphStorm accepts data in tabular format and provides a tool to construct a graph
based on the graph schema defined by users. After the graph construction,
a user applies GML methods to the data to train an initial model.
For model development,
GraphStorm provides end-to-end pipelines to train a model with
a single command line. To improve model performance, GraphStorm provides
many built-in techniques to tackle common problems in industry graph
applications; for more advanced users, GraphStorm provides
custom model APIs for trying brand new modeling ideas. Once the developed
model meets the deployment criteria, GraphStorm allows users to deploy
the model without any code modification on the ML service platforms,
such as Amazon SageMaker.

To cover a large number of industry applications, GraphStorm provides
modularized implementations to support diverse graph data and applications.
Figure \ref{fig:funcs} shows the abstractions required for model training,
including datasets, models, data loaders, tasks, training algorithms and evaluations.
GraphStorm currently supports four types of graph data: homogeneous graphs,
heterogeneous graphs, temporal graphs and textual graphs. For different types of graph
data, GraphStorm provides different modeling techniques, such as
RGCN \cite{rgcn} and HGT \cite{hgt} for heterogeneous graphs and TGAT \cite{tgat}
for temporal graphs.
GraphStorm provides task-specific data loaders: node data loaders for node-level tasks,
edge data loaders for edge-level tasks and graph data loaders for graph-level tasks.
GraphStorm explicitly provides separate data loaders for predicting an attribute of
an edge (\textit{EdgeDataLoader}) and for predicting the existence of an edge
(\textit{LinkPredictionDataLoader}) for the sake of efficiency because
\textit{LinkPredictionDataLoader} not only samples positive edges from a graph but also
adopts many different methods to construct negative edges for efficient training
(see Section \ref{sec:lp}).
GraphStorm supports different training strategies, such as
single-task training, multi-task training, LM-GNN co-training and Expectation–maximization (EM) training method. GraphStorm currently supports
seven graph tasks and provides corresponding evaluation metrics for model evaluation. By combining different components, we can have a complete solution
for a graph application. For example, as illustrated in
Figure \ref{fig:funcs}, if we need to perform
a link prediction task on a heterogeneous graph with rich text features,
we can take \textit{RGAT} as a graph model, use \textit{LinkPredictionDataLoader}
as the data loader and \textit{LM+GNN co-training} as the training algorithm
to train a model for \textit{link prediction}. We can use \textit{MRR} for
model evaluation.

\subsection{Architecture}
\begin{figure}[!htbp]
\centering
\includegraphics[width=0.5\textwidth]{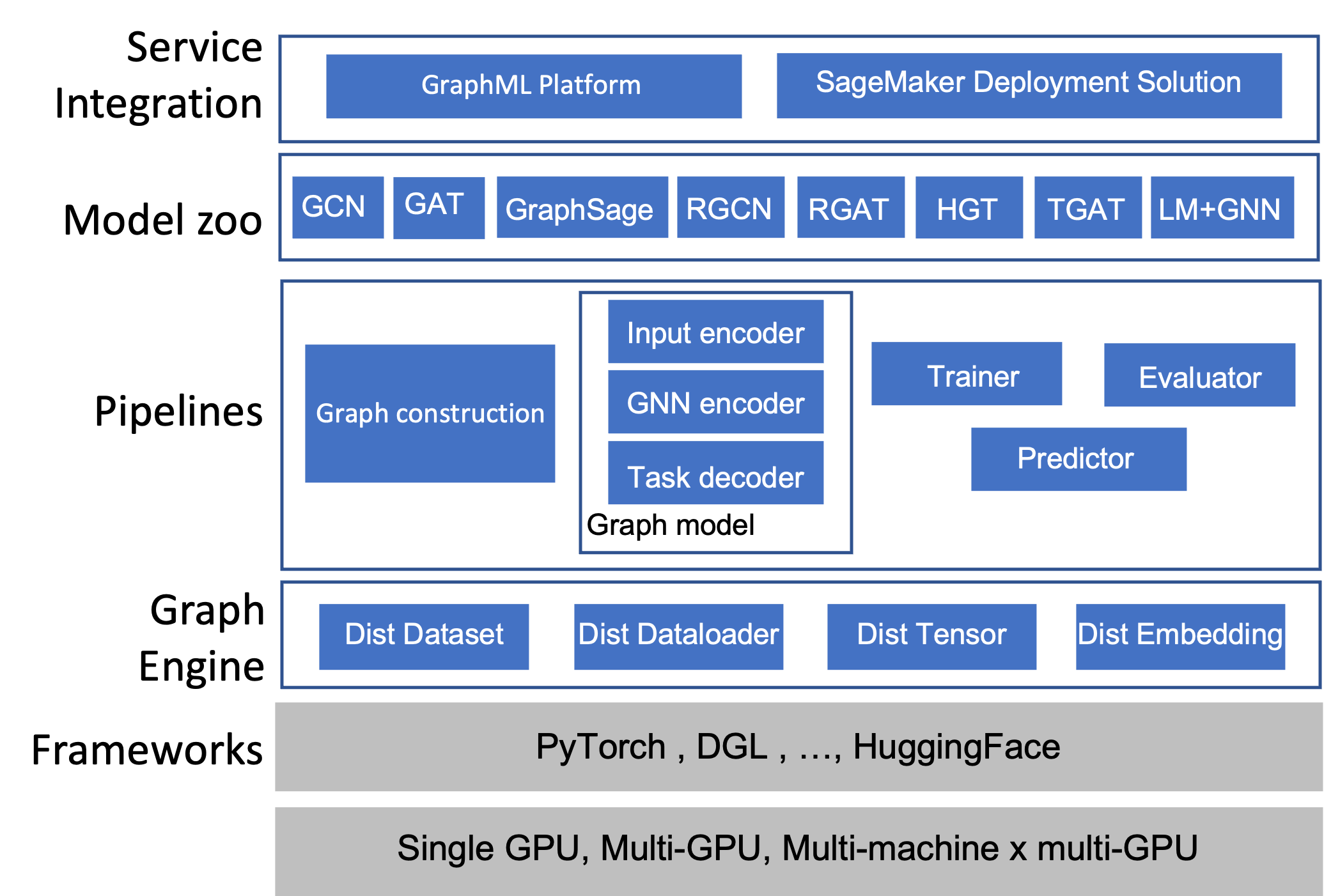}\\
\caption{GraphStorm architecture}
\label{fig:arch}
\end{figure}

To realize the modularized implementations, GraphStorm is designed with
four layers: the distributed graph engine, pipelines for graph construction,
training and inference, a general model zoo and service
integration, as illustrated in Figure \ref{fig:arch}. GraphStorm ensures that
every layer is scalable in order to achieve scalability.

\subsubsection{Distributed graph engine}
The first layer is the distributed graph engine built on top of DistDGL
\cite{distdgl}, which enables distributed training and inference on billion-scale graphs.
It provides the abstractions of datasets and data loaders to help users access
different types of graph data (e.g. homogeneous graphs and heterogeneous graphs)
stored in different hardware (e.g., in CPU memory or distributed memory).
It provides distributed tensors and distributed embedding layers
to help users store and access node/edge features and learnable node embeddings
in a cluster of machines.
The distributed graph engine provides the same interface on different hardware to
hide the complexity of handling different hardware.
GraphStorm adopts on-the-fly mini-batch sampling on a distributed graph.
The benefit of on-the-fly sampling is to allow sampling mini-batch data
with different configurations without preprocessing the input data again.
This is particularly useful for tuning hyperparameters, such as the batch size
and the number of GNN layers.

\subsubsection{Graph construction pipeline}
\vspace*{5mm}
Enterprise data are usually stored in RDBMS and NoSQL data stores. None of
them can be directly used by GML frameworks to conduct ML tasks.
GraphStorm provides a graph processing pipeline that takes tabular data as input
and constructs a graph for model training and inference.
GraphStorm provides a single-machine implementation to enable easly model prototyping
and a distributed Spark-based \cite{spark} implementation for scalable deployment.
Both implementations take the input data and output graphs in the same
format, making it easy for users to move from small-scale experiments to large-scale deployments.

The graph construction pipeline takes multiple steps to transform data
and eventually converts them into a graph format.
The first step
is feature transformation. To stay within a single library
for data processing, model training and inference,
GraphStorm provides implementations of many feature encoding methods that
accept numerical, categorical and text data and apply appropriate transformations
on the features at a billion scale in a distributed manner.
If the input data use strings to identify nodes and edges, GraphStorm also
converts them into integers because
GraphStorm model training and inference requires nodes and edges are identified
by integers. GraphStorm provides a distributed ID mapping mechanism that creates
massive mapping tables from strings to integers and applies them to all string IDs
on nodes and edges.
After feature transformation and ID mapping, GraphStorm provides a distributed
graph partitioning framework with various edge-cut algorithms, such as
random and METIS partitioning \cite{karypis2011parmetis, karypis1998metis},
which assigns nodes and edges to partitions.
The graph partititon algorithms in GraphStorm support
massive graphs with billions of nodes and edges.
The implementation of graph partitioning algorithm is decoupled from other
components, making it straightforward to introduce new distributed partitioning algorithms.
After the graph partition algorithm, the pipeline shuffles node data and edge
data in a distributed manner, constructs graph partition objects on each
machine in parallel and saves them in the DistDGL format, which is used by
GraphStorm for model training and inference.


\subsubsection{Training and inference pipeline}
The training/inference pipeline exposes GML model templates and supports different
graph tasks (node-level predictions, edge-level predictions and graph-level
predictions).
GraphStorm splits a GML model into three components: node/edge input
encoders, graph encoders and task decoders.
The node/edge input encoders handle
node/edge features (e.g., compute BERT embeddings from text features or read
learnable embeddings on featureless nodes).
The graph encoders use GNN to encode
the graph structure and node/edge features together to generate embeddings of nodes.
The task decoder is usually designed specifically for a task.
The pipelines provide trainers, predictors, and evaluators to train, infer and
evaluate a task-specific GML model that is implemented with the graph model interface
(shown in Section \ref{sec:api}). Because the models in the GraphStorm model zoo
are implemented with the same model interface, the same pipelines are used to
train/infer/evaluate GraphStorm's built-in models as well as users' custom models.


\subsubsection{Model zoo}
To lower the effort of developing GML models and improving
performance, GraphStorm provides a model zoo with a diverse
set of GML model implementations. This includes
RGCN \cite{rgcn}, RGAT \cite{rgat} and HGT \cite{hgt} for heterogeneous graphs, GCN \cite{gcn}, GAT \cite{gat} and GraphSage \cite{graphsage} for homogeneous graphs,
TGAT \cite{tgat} for temporal graphs, LM+GNN \cite{bertgnn} for text-rich graphs.
These model implementations can be used with different task decoders to handle different
graph tasks.

\subsection{User interface}

GraphStorm is designed to lower the bar of developing GML models for beginner users
as well as supporting advanced GML scientists to experiment with new methods on
industry-scale graph data. To support these diverse requirements, GraphStorm
provides two types of interfaces.
The command line interface allows users to quickly prototype a GML model on
their application data and allows them to improve model performance with builtin
GML techniques in GraphStorm. The programming interface allows advanced users
to develop their custom models to further improve model performance for their particular
applications.

\subsubsection{Command line interface}
\vspace*{5mm}

GraphStorm provides the command-line interface for graph construction,
model training, performance tuning and model inference.
For graph construction, GraphStorm provides the 
$gconstruct.construct\_graph$ module that takes the application data stored in
tabular format (e.g., CSV and Parquet) and constructs a graph with the GraphStorm
format for training and inference based on the graph schema defined by users.
For model development, GraphStorm provides a module for each graph task.
For example, GraphStorm provides $run.gs\_node\_classification$ to
train and run inference of node classification models. GraphStorm allows users to
configure model training and inference to enable many built-in techniques
(shown in Section \ref{sec:tech})
by using the command arguments. Appendix \ref{command} shows examples of
GraphStorm commands used for graph construction, model training and inference.

\subsubsection{Programming interface} \label{sec:api}
\vspace*{5mm}

GraphStorm provides the programming API to help advanced users implement
their custom models to further improve model performance. GraphStorm provides
task-specific trainers, predictors, evaluators as well as model templates.
Figure \ref{fig:node_model} shows a GraphStorm training script to train
an RGCN node classification model for illustration. At a minimum,
a user only needs 8 lines of code in the training script to train and
evaluate a model.
Appendix \ref{app:api} shows an example of defining a graph model with
GraphStorm's custom model API. By using GraphStorm's programming API,
a user only needs to focus on model development and GraphStorm handles
scaling the model training and inference because the same training
script can run on a single
GPU, multiple GPUs and multiple machines without modifying any code.
This greatly reduces the effort of prototyping models on the subset of an application
data and scaling the model training and deployment on the actual data.

\begin{figure}[!htbp]
\centering
\begin{lstlisting}[language=Python]
import graphstorm as gs

gs.initialize()
data = gs.dataloading.GSgnnData(part_config_file,
	node_feat_field='feat',
	label_field='label')
model = RGCNNCModel(train_data.g,
	num_gnn_layers=2, num_hidden=128, num_classes=2)
evaluator = gs.eval.GSgnnAccEvaluator(multilabel=False)
dataloader = gs.dataloading.GSgnnNodeDataLoader(data,
	train_idxs, fanout=[5, 5], batch_size=1024)
val_dataloader = gs.dataloading.GSgnnNodeDataLoader(data,
	eval_idxs, fanout=[5, 5], batch_size=1024)
trainer = gs.trainer.GSgnnNodeTrainer(model,
	eval=evaluator)
trainer.fit(train_dataloader=dataloader,
	val_dataloader=val_dataloader,
	num_epochs=10)
\end{lstlisting}
\vspace{-2mm}
\caption{GraphStorm training script for node classification.}
\label{fig:node_model}
\end{figure}

\subsection{Modeling techniques} \label{sec:tech}

Industry graphs are usually heterogeneous with multiple node types and edge types.
There are many modeling challenges associated with them. Below are a few common problems that we encounter when modeling industry graphs.
\begin{itemize}
    \item Some nodes/edges are associated with multi-modal features, such as text
    and images. How to jointly model text/image data with graph data together?
    \item Some node types do not have node features. How to model these nodes
    efficiently and effectively?
    \item It is common that some nodes in a dataset have no neighbors or
    few neighbors. How to improve the model performance on these isolated nodes
    with graph modeling?
    \item Link prediction is a common task in industry applications. However,
    the training set of link prediction is usually large (billions of edges).
    How to effectively and efficiently train a GML model with link prediction
    on a billion-scale graph?
\end{itemize}

GraphStorm provides built-in techniques to address these modeling issues to
improve model performance. GraphStorm only adopts
methods that can scale to billion-scale graphs and also ensures that
their implementations are scalable.

\subsubsection{Joint modeling text and graph data}
\vspace*{5mm}

Many industry graph data have rich text features on nodes and edges. For example,
the Amazon search graph has rich text features on queries (keywords) and products
(product descriptions) \cite{xie2023graph}. This requires jointly modeling graph and text data together.
By default, GraphStorm runs language models (LM), such as BERT, on the text features
and runs a GNN model on the graph structure in a cascading manner.

GraphStorm implements some efficient methods to train LM models and GNN models for
downstream tasks \cite{bertgnn, glem}. For example, Ioannidis et al. \cite{bertgnn}
devises a three-stage training method that first performs graph-aware fine-tuning
of the language model, train the GNN model with the fixed language model and
follow with the end-to-end fine-tuning; GLEM \cite{glem} trains the LM model and
the GNN model iteratively on the downstream application. The original GELM paper
was designed for homogeneous graphs and GraphStorm extends it to heterogeneous graphs.

\subsubsection{Efficient training large embedding tables for featureless nodes}
\vspace*{5mm}

In an industry graph, some node types do not have any node features. For example,
when constructing the Microsoft academic graph (MAG), it is hard
to construct node features on author nodes; as a result, we typically do not have
any features on the author nodes when constructing the graph. When training a GNN model
on the graph, GraphStorm
by default adds learnable embeddings on author nodes, which results in a massive
learnable embedding table.
There are two problems of using learnable embeddings for featureless nodes.
First, the learnable embedding table is usually massive because a graph can have many
featureless nodes (e.g., there are 200 million authors in the MAG graph). Secondly,
adding learnable embeddings significantly increases the model size, which easily causes
overfitting.
GraphStorm provides additional options to handle featureless nodes.

GraphStorm provides methods that construct node features of the featureless nodes
with the neighbors that have features.
\begin{equation}
F'_{v_{i}} = f(F_{v_{j}}, \forall j \in N_{i})
\end{equation}
\noindent With the constructed node features, a user can use a normal GNN model to
generate GNN embeddings. We can have different choices for $f$. We can use
non-learnable function (e.g., average) or learnable functions (e.g., Transformer)
to construct node features.

GraphStorm also provides a two-stage training method to train
learnable embeddings and GNN models, similar to the multi-stage training of LM+GNN.
In the first stage, we use link prediction to train the learnable embeddings.
In the second stage, we fix the learnable embeddings as node features and train
a GNN model for downstream tasks.




\subsubsection{GNN distillation for isolated nodes}
\vspace*{5mm}

In many industry graphs, there exist nodes that do not have neighbors, known
as \textit{isolated} nodes. For example, in e-commerce applications, a search graph is constructed from customer logs over some period of time (e.g., purchases and clicks connecting the search queries and the products) for model training. When the model is deployed, it is common that new products and queries are added. These new nodes are not covered by the graph during model training and sometimes
do not connect with any nodes in the graph.
Simply applying the trained GNN model on these isolated nodes is not effective. 
To address this issue, GraphStorm provides the capability of distilling
a GNN model into another model without graph dependency (e.g., MLP or DistilBERT \cite{sanh2019distilbert}) \cite{zheng2022cold, zhang2022graphless}.
GraphStorm provides
multiple options for distillation (e.g., using soft labels or using embeddings directly).

\subsubsection{Optimizations for link prediction}\label{sec:lp}
\vspace*{5mm}
Link prediction is widely used in the industry as a pre-training method to produce high-quality entity representations. However, performing link prediction training on large graphs is non-trivial both in terms of model performance and efficiency. GraphStorm provides many techniques to optimize link prediction.

GraphStorm incorporates three ways of improving the model performance of link prediction.
Firstly, to avoid information leak in model training, GraphStorm by default excludes validation and testing edges from training graphs in the model training~\cite{zhu2023spottarget}. GraphStorm also excludes training target edges from message passing to avoid model overfitting.
Secondly, to better handle heterogeneous graphs, GraphStorm provides two ways to compute link prediction scores: dot product and DistMult~\cite{yang2015embedding}.
Dot product works when there is only one training edge type, while DistMult works when there are multiple training edge types.
Thirdly, GraphStorm provides two options to compute training losses, i.e., cross entropy loss, and contrastive loss. The cross entropy loss turns a link prediction task into a binary classification task. 
The contrastive loss compels the representations of connected nodes to be similar while forcing the representations of disconnected nodes to be dissimilar. 
Contrastive loss is more stable to different numbers of negative samples and converges faster.
In contrast, cross entropy loss only works with small number of negative samples and requires more time to converge. However, cross entropy loss allows users to define the importance, i.e., a weight, for each positive edge, providing extra flexibility when modeling positive samples.

Link prediction requires constructing numerous negative edges to train
a model. In distributed training, constructing negative edges naively may result
in accessing a large number of nodes from remote partitions. This would cause
a large amount of data movement between machines and slow training speed.
GraphStorm provides multiple negative sampling methods, such as uniform negative
sampling, joint negative sampling and in-batch negative sampling, to trade-off training efficiency and model performance. For example, uniform sampling may allow
the best model performance but requires a mini-batch to involve numerous
nodes from remote partitions, which results in slow training speed; in-batch
negative sampling only uses the nodes within the mini-batch to construct negative
edges, which minimize the number of nodes in a mini-batch to speed up training,
but may have some model performance degradation. GraphStorm provides all of these
options to give users more options to train their models (see details in Appendix~\ref{appendix-lp}).

\section{Evaluation}
\vspace*{5mm}

\subsection{Datasets} \label{sec:data}

\begin{table*}[t]
\centering
\caption{The statistics of benchmark datasets.}
\label{tab:data}
\begin{tabular}{ c  c  c  c  c  c  c}
 \toprule
 Dataset & \#nodes & \#edges & \#node/edge types & NC training set & LP training set & text-feature nodes  \\
 \midrule
 Amazon Review & 286,462,374 & 1,053,940,310 & 3/4 & 1,826,784 & 25,984,064 & 242,967,461\\
 MAG & 484,511,504  & 7,520,311,838 & 4/4 & 28,679,392 & 1,313,781,772 &240,955,156\\
\bottomrule
\end{tabular}
\end{table*}

We evaluate GraphStorm with two large heterogeneous graphs: Microsoft academic
graph \cite{sinha2015overview} and Amazon review graph \cite{mcauley2015image} (shown in Table~\ref{tab:data}). We construct two tasks on the benchmark datasets.
For node classification, we predict the paper venue on the MAG data and predict
the brand of a product on the Amazon review dataset; for link prediction, we predict
what papers a paper cites on the MAG dataset and predict which products are
co-purchased with a given product on the Amazon review dataset.
More details of the benchmark datasets can be found in the Appendix.


\subsection{GML on Industrial-Scale Graphs}

We benchmark two main methods (pre-trained BERT+GNN and fine-tuned BERT+GNN) in
GraphStorm on the two large datasets. For pre-trained
BERT+GNN, we first use the pre-trained BERT model to compute BERT embeddings from
node text features and train a GNN model for prediction.
The alternative solution is to fine-tune the BERT models on the graph data before
using the fine-tuned BERT model to compute BERT embeddings
and train GNN models for prediction. We adopt different ways to fine-tune the BERT
models for different tasks. For node classification, we fine-tune the BERT model
on the training set with the node classification tasks; for link prediction, we
fine-tune the BERT model with the link prediction tasks.

\begin{table*}[h]
\centering
\caption{The overall performance and computation time of GraphStorm. 
}
\label{tbl:overall}
\scalebox{0.9}{
\begin{tabular}{ cccc ccccc c}
 \toprule
 \multirow{2}{*}{\textbf{Dataset}}  & \multirow{2}{*}{\textbf{Task}} & \multirow{2}{*}{\textbf{Data process}} &\multirow{2}{*}{\textbf{Target}} & \multicolumn{3}{c}{\textbf{Pre-trained BERT + GNN}} &  \multicolumn{3}{c}{\textbf{Fine-tuned BERT + GNN}}\\
  &  &  & & LM Time Cost & Epoch Duration & Metric &  LM Time Cost & Epoch Duration & Metric\\
 \midrule
 \multirow{2}{*}{\textbf{MAG}} & NC & \multirow{2}{*}{9:13:00} & venue & 3:26:00 & 2:14:33	& Acc:0.5715 & 23:42:35	& 2:16:42 & Acc: 0.6333 \\
 & LP &  & cite & 3:18:19 & 36:34:52 & Mrr: 0.4873 & 75:08:22 & 36:12:17 & Mrr: 0.6841\\
 \multirow{2}{*}{\textbf{AR}} & NC & \multirow{2}{*}{3:35:00} & brand & 8:03:23	& 0:02:47 & Acc: 0.8407 & 8:09:54 & 0:02:48 & Acc: 0.8963 \\
 & LP & & co-purchase & 7:28:12 &	0:44:12	& MRR: 0.9602 & 7:42:18& 0:33:19 & MRR: 0.9710 \\
\bottomrule
\end{tabular}
}
\end{table*}

Table \ref{tbl:overall} shows the overall model performance of the two methods and
the overall computation time of the whole pipeline starting from data processing
and graph construction.
In this experiment, we use 8 r5.24xlarge instances for data processing and use
4 g5.48xlarge instances for model training and inference.
Overall, GraphStorm enables efficient graph construction and model
training on large text-rich graphs with hundreds of millions of nodes.
We can see that for most cases, GraphStorm can finish all steps, even
the expensive BERT computations, within a few hours. The only exception is
fine-tuning BERT model with link prediction and training a GNN link prediction model
on the MAG dataset, which take 2-3 days. The reason is that the training set of
the dataset for link prediction has billions of edges and we fine-tune BERT and
train GNN model on the entire training set.
For pre-trained BERT+GNN, it takes 3.5 hours to compute BERT embeddings on
240 million paper nodes in MAG and takes 8 hours on 240 million nodes in
the Amazon review data.
To improve model performance, we fine-tune the BERT
model on the training set of the downstream tasks, which leads to a significant
performance boost. Its performance improvement is also significant. For example,
fine-tuning BERT on MAG can improve BERT+GNN by 11\% for node classification
and by 40\% for link prediction.

\begin{table}[t]
\centering
\caption{Scalability of GraphStorm on synthetic graphs with 1B, 10B and 100B edges. The corresponding training set sizes are 8 million, 80 million and 800 million, respectively.}
\label{tab:100b}
\scalebox{0.85}{
\begin{tabular}{ c  c  c  c c c c}
 \toprule
 Graph & \multicolumn{2}{c}{Data pre-process} & \multicolumn{2}{c}{Graph Partition} & \multicolumn{2}{c}{Model Training} \\
 Size & \# instances & Time & \# instances & Time & \# instances & Time \\
 \midrule
 1B & 4 & 19 min & 8 & 8 min & 8 & 1.5 min\\
 10B & 8  & 31 min & 16 & 41 min &  16 & 8 min \\
 100B & 16  & 61 min & 32 & 416 min & 32 & 50 min \\
\bottomrule
\end{tabular}
}
\end{table}

We also benchmark GraphStorm on large synthetic graphs to further illustrate
its scalability. We generate three synthetic graphs with one billion, 10 billion
and 100 billion edges respectively. The average degree of each graph is 100 and
the node feature dimension is 64. We train a GCN model for node classification
on each graph. In this experiment, we use r5.24xlarges instances for data
pre-processing, graph partition and model training. For graph partition, we use
a random partition instead of METIS partition. For model training, we take
80\% of nodes as training nodes and run the training for 10 epochs. Table~\ref{tab:100b} shows the computation time of graph pre-processing,
graph partition and model training. Overall, GraphStorm enables graph
construction and model training on 100 billion scale graphs within hours and shows good scalability. For data pre-processing, when increasing the graph size from 1 billion to 100 billion edges, the overall computation cost (measured by instance-minutes) only increases by $13\times$.
For model training, when increasing both the graph size and training set by 100X, the overall training cost only increases by $133\times$. For graph partition, when increasing the graph size by $100\times$, the overall computation cost only increases by $208\times$.

\subsection{Modeling graph data from data preprocessing}

\begin{table}[t]
\centering
\caption{Performance on Amazon Review Graph varying graph schemas.}
\label{tab:vary-schema}
\scalebox{0.9}{
\begin{tabular}{ c  c  c  c c}
 \toprule
 Schema & node types & featureless & LP & NC \\
 \midrule
 Homogenous & item & No & MRR:0.937 & Acc:0.640 \\
 Heterogenous-v1 & +review  & No & MRR:0.944 & Acc:0.742 \\
 Heterogenous-v2 & +customer  & ``customer'' & MRR:0.960	& Acc:0.725 \\
\bottomrule
\end{tabular}
}
\end{table}

GML leverages the inductive bias inherent in graph structures, making the graph schema crucial for the successful application of GML. GraphStorm offers an efficient
graph construction pipeline to power scientists to experiment GML modeling starting from the graph construction stage. We use the Amazon Review graph to illustrate
the importance of graph schema. We gradually increase the heterogeneity of the graph schema and observe the corresponding variations in model performance.
As illustrated in Table~\ref{tab:vary-schema}, adding review nodes and the (item, \textit{receives}, review) edge enhances performance in both co-purchase prediction and brand classification tasks. Moreover, incorporating featureless customer nodes and the (customer, \textit{writes}, review) edge further improves co-purchase prediction performance, but not node classification. This improvement is likely because items reviewed by the same customer have a higher likelihood of being purchased together, although this factor does not significantly influence the determination of an item's brand. This indicates the importance of defining
the right graph schema to improve performance and that users need to try out different graph schemas to determine
the best schema for a given task.

\subsection{Performance of GraphStorm techniques}
Besides the aforementioned GNN and LM+GNN models, GraphStorm supports many techniques
to improve model performance on industry graph data. In this section,
we evaluate some of the techniques described in Section \ref{sec:tech}
on our benchmark datasets.

\subsubsection{Joint text and graph modeling}

\begin{figure}[h]
\centering
\includegraphics[width=0.8\linewidth]{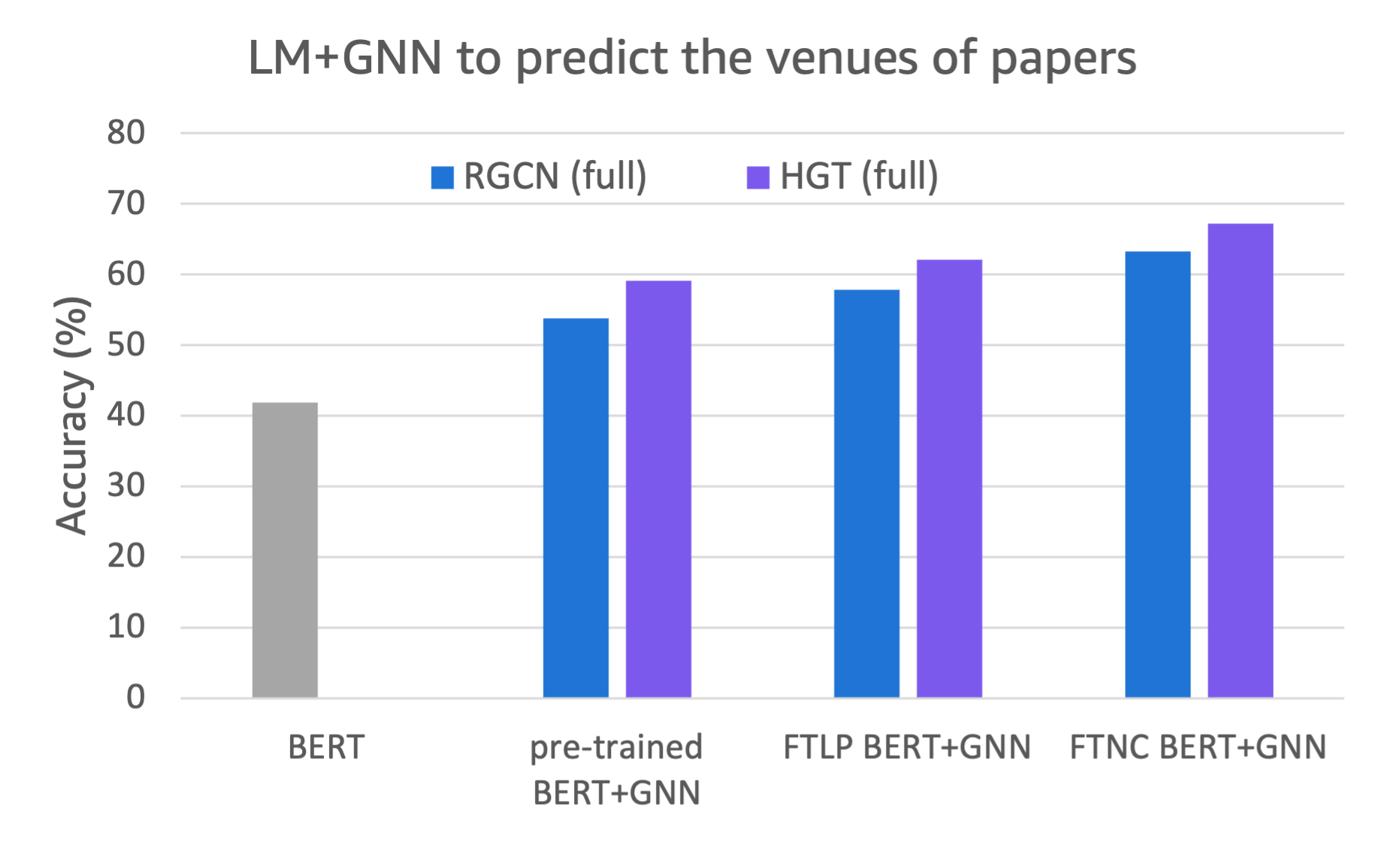}\\
\vspace{-2mm}
\caption{Jointly modeling text and graph data on Microsoft
Academic Graphs}
\label{fig:joint-modeling}
\end{figure}

We benchmark GraphStorm's capability of jointly modeling text and graph data. We use
different methods to train BERT and GNN on the full MAG data to illustrate
the effectiveness of different methods (Figure~\ref{fig:joint-modeling}). Here we
compare four different methods: fine-tune BERT to predict the venue, train GNN with
BERT embeddings from pre-trained BERT model (pre-trained BERT+GNN), train GNN
with BERT embeddings generated from a BERT model fine-tuned with link prediction
(FTLP BERT+GNN), train GNN with BERT embeddings generated from a BERT model
fine-tuned with venue prediction (FTNC BERT+GNN).
First, BERT+GNN is much more effective in predicting paper venues than
BERT alone by up to 54\%. It also shows the importance of BERT fine-tuning in
the prediction task. Even though fine-tuning BERT with link prediction can
improve the prediction accuracy (by up to 7.6\% over pre-trained BERT+GNN),
the best accuracy is achieved by first fine-tuning
the BERT model with venue prediction and training GNN with the fine-tuned BERT model
with venue prediction (by 17.6\%).

\subsubsection{GNN distillation}
\vspace*{5mm}
\begin{table}[t]
\centering
\caption{Performance of GNN Embeddings Distillation on MAG Dataset.}
\label{tbl:distil}
\begin{tabular}{  c  c  c  c  c  c }
 \toprule
 Settings & Acc \\
 \midrule
 DistilBERT fine-tuned with venue labels (768 dim) & 41.17\% \\
 \hline
 DistilBERT with GNN distillation (128 dim) & 44.53\% \\
 \bottomrule
\end{tabular}

\end{table}

We evaluate the GNN distillation capability in GraphStorm to improve the performance
of the BERT-like language models. In this experiment, we train a GNN model to
predict venues of papers in MAG and distill it to a language model; we compare this distilled model with
a language model as the baseline that is directly fine-tuned to predict venues.
We choose pre-trained 70MM HuggingFace DistilBERT \cite{sanh2019distilbert} for
the BERT model architecture and use its initial weights for both the baseline and
the GNN distilled model. For the baseline, we fine-tune the DistilBERT using
the venue labels. For GNN distilled model, we conduct
the distillation to minimize the distance between the embeddings from a GNN teacher model and the embeddings from a DistilBERT student model. MSE loss is used to supervise the training. After the baseline DistilBERT and GNN-distilled DistilBERT are trained, we use them to generate embeddings on paper nodes separately. We evaluate the performance of the generated embeddings by training MLP decoders for embeddings from baseline DistlBERT and GNN-distilled DistilBERT separately.
As shown in Table \ref{tbl:distil}, the GNN-distilled DistilBERT embeddings outperform the baseline DistilBERT embeddings by 8.2\%, demonstrating the structured knowledge from the GNN teacher model are transferred to the GNN-distilled DistilBERT.

\subsubsection{Link prediction}
We benchmark GraphStorm's capability of training link prediction models on Amazon Review dataset. We conduct link prediction on the (item, also-buy, item) edges. We compare the model performance and training time with different loss functions and negative sampling settings. The loss functions include contrastive loss and cross entropy loss.
The negative sampling methods include in-batch negative sampling, joint negative sampling with the number of negatives of 1024, 32 and 4 (denoted as joint-1024, joint-32 and joint-4, respectively), and uniform negative sampling with the number of negatives of 1024 and 32 (denoted as uniform-1024 and uniform-32, respectively). We set the local batch size to 1024 and the maximum training epochs to 20.

Table~\ref{tab:lp-result} shows the result. 
Overall, the performance of contrastive loss is much better than cross entropy loss. Contrastive loss is more robust to the variance of the number of negative edges. Cross entropy loss works much better when the number of negatives is small, e.g., 4, but its performance becomes lower when the number of negatives is larger. Furthermore, contrastive loss converges much faster than cross entropy loss. 
In general, uniform negative sampling is more computationally expensive and consumes more GPU memory than the other two sampling methods, as it samples many more negative nodes. For example, with batch size of 1024, both in-batch and joint-32 samples 1024 nodes to construct negative edges, while uniform-32 has to sample 32,768 nodes. So the epoch time of uniform sampling is larger than the other two.

\begin{table}[t]
\centering
\caption{Performance of link prediction on Amazon Review Graph with varying training techniques.}
\label{tab:lp-result}
\scalebox{0.9}{
\begin{tabular}{ c  c  c  c c}
 \toprule
 Loss func & Neg-Sample & epoch time & \#epochs & Metric \\
 \midrule
 contrastive & in-batch & 1340.90s & 5   & MRR:0.951 \\
contrastive & joint-1024  & 1344.65s & 8 & MRR:0.956 \\
contrastive & joint-32  & 1286.64s & 8 & MRR:0.958 \\
contrastive & joint-4  &  1289.9s  & 10 &  MRR:0.956 \\
contrastive & uniform-32  & 1726.19s & 8 & MRR: 0.957 \\
contrastive & uniform-1024  & \multicolumn{3}{c}{OOM} \\
\midrule
cross-entropy & in-batch  & 1343.94s & 20& MRR: 0.250 \\
cross-entropy & joint-1024  & 1330.50s & 18	& MRR:0.334 \\
cross-entropy & joint-32  & 1290.72s & 15	& MRR:0.380 \\
cross-entropy & joint-4  & 1288.53s & 16 & MRR:0.645 \\
cross-entropy & uniform-32  & 1746.68s & 20	& MRR:0.377 \\
cross-entropy & uniform-1024  & \multicolumn{3}{c}{OOM} \\
\bottomrule
\end{tabular}
}
\end{table}

\section{Discussion and Conclusions}
GraphStorm is a general no-code/low-code GML framework designed for industry
applications. It provides end-to-end pipelines for graph construction, model training
and inference for many different graph tasks and scales to industry graphs with
billions of nodes efficiently.
This helps scientists develop GML models starting from graph schema definition and
GML model prototyping on large industry-scale graphs without writing code. Based on
our experience working with scientists in the industry, this significantly reduces the overhead
of developing a new GML model and tuning its model performance for
an industry application. GraphStorm provides many advanced GML modeling techniques
to handle problems commonly encountered in industry applications and
we have fully verified the effectiveness and scalability of the techniques on
the public and private benchmarks. By using GraphStorm, ML scientists
can quickly prototype GML models and use built-in techniques to improve model
performance and outperform production models.
In addition to using the built-in techniques
to improve model performance, users can also develop their own custom models in
GraphStorm and rely on GraphStorm to scale their model training to industry-scale
graphs in a cluster of machines.
Currently, GraphStorm has been used to develop and deploy GML models for over
a dozen industry applications. We believe that the capability provided by GraphStorm
can also help GML researchers in the research community to experiment
new modeling techniques on large complex graph data.



%
\newpage
\bibliographystyle{abbrv}
\bibliography{gsf}  
%
%
\clearpage
\newpage
\appendix

\section{Link prediction score functions and loss functions} \label{appendix-lp}
\subsection{Score functions} GraphStorm provides two score functions:
\begin{itemize}
    \item \textbf{Dot Product}:
    \begin{equation}
    score(v_i, v_j) = \sum_{i=0}^{n-1} {emb_{v_i} * emb_{v_j}}
    \end{equation}
    where $v_i$ and $v_j$ are two nodes, $emb_{v_i}$ is the embedding of $v_i$, $emb_{v_j}$ is the embedding of $v_j$, and the dimension of $emb_{v_i}$ and $emb_{v_j}$ is $n$.

    \item \textbf{DistMult}:
    \begin{equation}
    score(v_i, v_j) = \sum_{i=0}^{n-1} {emb_{v_i} * emb_{rel_{ij}} * emb_{v_j}}
    \end{equation}
    where where $v_i$ and $v_j$ are two nodes, $emb_{v_i}$ is the embedding of $v_i$, $emb_{v_j}$ is the embedding of $v_j$, $emb_{rel_ij}$ is the embedding regarding to the edge type of the edge between $v_i$ and $v_j$, and the dimension of $emb_{v_i}$ and $emb_{v_j}$ is $n$.
\end{itemize}

\subsection{Loss functions} 
GraphStorm provides three loss functions:
\begin{itemize}
    \item \textbf{Cross entropy}: The cross entropy loss turns a link prediction task into a binary classification task. We treat positive edges as 1 and negative edges as 0. The loss of edge $e$ is as following:
    \begin{equation}
        loss = - y \cdot \log score + (1 - y) \cdot \log (1 - score)
    \end{equation}
    where $y$ is 1 when $e$ is a positive edge and 0 when it is a negative edge. $score$ is the score number of $e$ computed by the score function.
    
    \item \textbf{Weighted cross entropy}: The weighted cross entropy loss is similar to \textbf{Cross entropy} loss except that it allows users to set a weight for each positive edge. 
    The loss function of $e$ is as following:
    \begin{align}
        loss = - w_{e} \left[ y \cdot \log score + (1 - y) \cdot \log (1 - score) \right]
    \end{align}
    where $y$ is 1 when $e$ is a positive edge and 0 when it is a negative edge. $score$ is the score number of $e$ computed by the score function, $w_{e}$ is the weight of $e$ and is defined as
    \begin{equation}
     w_{e} = \left \{ 
     \begin{array}{lc}
         1,  & \text{ if } e \in G, \\
         0,  & \text{ if } e \notin G
     \end{array}
     \right.
    \end{equation}
    where $G$ is the training graph.

    \item \textbf{Contrastive loss}: The contrastive loss compels the representations of connected nodes to be similar, while forcing the representations of disconnected nodes remain dissimilar. In the implementation, we use the score computed by the score function to represent the distance between nodes. When computing the loss, we group one positive edge with the $N$ negative edges corresponding to it.
    The loss function is as following:

    \begin{equation}
    loss = -log(\dfrac{exp(pos\_score)}{\sum_{1=0}^N exp(score_i)})
    \end{equation}
    where $pos\_score$ is the score of the positive edge. $score_i$ is the score of the i-th edge. In total, there are $N+1$ edges, within which there is 1 positive edge and N negative edges.

\end{itemize}

\subsubsection{Negative sampling methods} GraphStorm provides four negative sampling methods:

\begin{itemize}
    \item \textbf{Uniform negative sampling}: Given $N$ training edges under edge type $(src\_t, rel\_t, dst\_t)$ and the number of negatives set to $K$, uniform negative sampling randomly samples $K$ nodes from $dst\_t$ for each training edge. It corrupts the training edge to form $K$ negative edges by replacing its destination node with sampled negative nodes. In total, it will sample $N * K$ negative nodes.

    \item \textbf{Joint negative sampling}: Given $N$ training edges under edge type $(src\_t, rel\_t, dst\_t)$ and the number of negatives set to $K$, joint negative sampling randomly samples $K$ nodes from $dst\_t$ for every $K$ training edges. For these $K$ training edges, it corrupts each edge to form $K$ negative edges by replacing its destination node with the same set of negative nodes. In total, it only needs to sample $N$ negative nodes. (We suppose $N$ is dividable by $K$ for simplicity.)
    
    \item \textbf{Local joint negative sampling}: Local joint negative sampling samples negative edges in the same way as joint negative sampling except that all the negative nodes are sampled from the local graph partition.

    \item \textbf{In-batch negative sampling}: In-batch negative sampling creates negative edges by exchanging destination nodes between training edges. For example, suppose there are three training edges $(u_1, v_1)$, $(u_2, v_2)$, $(u_3, v_3)$, In-batch negative sampling will create two negative edges $(u_1, v_2)$ and $(u_1, v_3)$ for $(u_1, v_1)$, two negative edges $(u_2, v_1)$ and $(u_2, v_3)$ for $(u_2, v_2)$ and two negative edges $(u_3, v_1)$ and $(u_3, v_2)$ for $(u_3, v_3)$. If the batch size is smaller than the number of negatives, either of the above three negative sampling methods can be used to sample extra negative edges.

\end{itemize}

\section{Command line interface} \label{command}
\vspace*{5mm}

GraphStorm provides two options for graph construction:
\begin{itemize}
    \item \textit{graphstorm.gconstruct.construct\_graph} runs in a single machine. It is mainly used by the model developers to prototype a GML model on a subset of the application data.
    \item GraphStorm Distributed Data Processing (GSProcessing) runs on multiple machines. It is mainly designed for industry-scale deployment.
\end{itemize}
\noindent The command below shows how to construct a graph with
\textit{graphstorm.gconstruct.construct\_graph}. It takes the input data
and the graph schema defined by a user in a JSON file.

\begin{lstlisting}
python3 -m graphstorm.gconstruct.construct_graph \ 
    --num-processes NUM_PROCESSES \ 
    --output-dir OUTPUT_DIR_PATH \ 
    --graph-name GRAPH_NAME \
    --num-partitions NUM_PARTITIONS \
    --conf-file GRAPH_SCHEMA_JSON_FILE
\end{lstlisting}

Figure~\ref{sup:example_config} shows an example of the graph schema configuration in a JSON file that contains the descriptions of the information of nodes and edges.

\begin{figure}[]
\centering
\begin{lstlisting}
{
    "version": "gconstruct-v0.1",
    "nodes": [
        ......
        {
            "node_type": "paper",
            "format": {
                "name": "parquet"
            },
            "files": [
                "nodes/paper.parquet"
            ],
            "node_id_col": "node_id",
            "features": [
                {
                    "feature_col": "feat",
                    "feature_name": "feat"
                }
            ],
            "labels": [
                {
                    "label_col": "label",
                    "task_type": "classification",
                    "split_pct": [0.8, 0.1, 0.1]
                }
            ]
        },
        ......
    ],
    "edges": [
        ......
        {
            "relation": ["paper", "citing", "paper"],
            "format": {
                "name": "parquet"
            },
            "files": [
                "edges/paper_citing_paper.parquet"
            ],
            "source_id_col": "source_id",
            "dest_id_col": "dest_id",
            "labels": [
                {
                    "task_type": "link_prediction",
                    "split_pct": [0.8, 0.1, 0.1]
                }
            ]
        },
    ......
    ]
}
\end{lstlisting}
\caption{A graph schema JSON file example.}
\label{sup:example_config}
\end{figure}

GraphStorm provides different modules for different graph tasks. For example, GraphStorm
provides \textit{graphstorm.run.gs\_link\_prediction} for link prediction and provides
\textit{graphstorm.run.gs\_node\_classification} for node classification. The same module
can be used for both training and inference.

\begin{lstlisting}
python3 -m graphstorm.run.gs_link_prediction \
        --num-trainers NUM_TRAINERS \
        --part-config GRAPH_PARTITION_JSON \
        --ip-config IP_LIST_FILE \
        --cf MODEL_CONF_FILE \
        --save-model-path MODEL_OUTPUT_PATH
\end{lstlisting}

The main difference between training and inference is that users need to provide
$--inference$ to turn the module to the inference mode. It also requires
the argument $--restore-model-path$ to load the saved model artifacts and
the argument $--save-embed-path$ to save the node embeddings.

\begin{lstlisting}
python3 -m graphstorm.run.gs_link_prediction \
        --inference 
        --num_trainers NUM_TRAINERS \
        --part-config GRAPH_PARTITION_JSON \
        --ip-config IP_LIST_FILE \
        --cf MODEL_CONF_FILE \
        --save-embed-path EMBED_OUTPUT_PATH \
        --restore-model-path MODEL_PATH
\end{lstlisting}

\newpage
\section{Programming interface}\label{app:api}
To use GraphStorm's programming interface, users need to
implement the forward function to define how to compute the loss function on
a mini-batch to train the model, the prediction function to define how to compute
the prediction results on each node/edge, the function to create an optimizer,
the function to save a model and the function to restore a model from the saved
checkpoint.

\begin{figure}[!htbp]
\centering
\begin{lstlisting}[language=python]
class GSgnnNodeModelBase:
    def forward(self, blocks, node_feats,
        edge_feats, labels, input_nodes)
    def predict(self, blocks, node_feats,
        edge_feats, input_nodes, return_proba):
        
    def create_optimizer(self):
    
    def restore_model(self, model_path):
    
    def save_model(self, model_path):
    
\end{lstlisting}
\caption{The model template for node prediction tasks.}
\label{fig:API}
\end{figure}

\subsection{Customized Model Example}
GraphStorm allows users to develop their own GML models by extending custom model APIs, and run these models in GraphStorm's training and inference pipelines. Here is an example of a customer model in the Graphstorm tutorial (\url{https://graphstorm.readthedocs.io/en/latest/advanced/own-models.html}).

\begin{figure}
\centering
\begin{lstlisting}[language=python]
class HGT(GSgnnNodeModelBase):
    def __init__(self, ......)
    ......
    # use GraphStorm loss function components
    self._loss_fn = ClassifyLossFunc(multilabel=False)

    def forward(self, blocks, node_feats, _, labels, _):
        h = node_feats
        for i in range(self.num_layers):
            h = self.gcs[i](blocks[i], h)
        for ntype, emb in h.items():
            h[ntype] = self.out(emb)
        preds = h[self.target_ntype]
        targets = labels[self.target_ntype]
        loss = self._loss_fn(preds, targets)
        return loss

\end{lstlisting}
\caption{Customized HGT model.}
\label{fig:customer_model}
\end{figure}
\end{document}